\documentclass[conference]{IEEEtran}
\usepackage{graphicx}
\usepackage{amsmath}
\usepackage{csquotes}
\usepackage{breqn}

\IEEEoverridecommandlockouts
\IEEEpubid{\makebox[\columnwidth]{978--1--1235--4/17\$31.00~\copyright~2017 IEEE \hfill} \hspace{\columnsep}\makebox[\columnwidth]{}}

\begin{document}
\title{Vector Learning for Cross Domain Representations}



\author{
\IEEEauthorblockN{Shagan Sah, Chi Zhang, Thang Nguyen, Dheeraj Kumar Peri, Ameya Shringi, Raymond Ptucha}
\IEEEauthorblockA{Rochester Institute of Technology, Rochester, NY 14623, USA}
}

\maketitle

\begin{abstract}
Recently, generative adversarial networks have gained a lot of popularity for image generation tasks. However, such models are associated with complex learning mechanisms and demand very large relevant datasets. This work borrows concepts from image and video captioning models to form an image generative framework. The model is trained in a similar fashion as recurrent captioning model and uses the learned weights for image generation. This is done in an inverse direction, where the input is a caption and the output is an image. The vector representation of the sentence and frames are extracted from an encoder-decoder model which is initially trained on similar sentence and image pairs. Our model conditions image generation on a natural language caption. We leverage a sequence-to-sequence model to generate synthetic captions that have the same meaning for having a robust image generation. One key advantage of our method is that the traditional image captioning datasets can be used for synthetic sentence paraphrases. Results indicate that images generated through multiple captions are better at capturing the semantic meaning of the family of captions.


\end{abstract}



\section{Introduction}
As few as five years ago, the automatic annotation of image and videos with natural language descriptions seemed quite distant.  Recent discoveries in convolutional and recurrent neural networks have led to unprecedented vision and language understanding such that automatic captioning is now commonplace. These discoveries have fueled the growth of new capabilities such as improved description of visual stimulus for the blind, advanced image and video search, and video summarization.

An important objective for computer vision and natural language processing research is to be able to represent both modalities of data in a common latent vector representation. Concepts that are similar, lie close together in this space, while dissimilar concepts lie far apart. This allows, for example, a keyword search of ``boat'' to not only return images of boats, but similar words, similar sentences, videos, and audio clips.  Much like the International Color Consortium's device independent profile connection space for color management \cite{international2010specification,pawle2002evolution}, a source independent vector connection space requires each new modality to only define a single transformation into and out of this reference space, rather than define a transformation to and from all other modalities.

Our work borrows from recent advances in vector representations \cite{kiros2015skip,le2014distributed,zhang2017semantic}, generative models  \cite{nguyen2016plug}, image/video captioning \cite{donahue2015long, xu2015show,venugopalan2015sequence} and machine translation \cite{sutskever2014sequence} frameworks to form a multi-modal common vector connection space.  We demonstrate the concept for both images and sentences, and show the extension to other modalities such as words, paragraphs, video, and audio is possible.  

The main contributions of this paper are: 1) Developing a robust source independent vector connection space between vision and language by conditioning image generation on multiple sentences; and 2) Integrating a language translation model for synthesizing artificial sentences with image generation.

The rest of this paper is organized as follows: Section \ref{sec:related} reviews relevant techniques. Section \ref{sec:method} presents the multiple caption conditioned image generation framework. Section \ref{sec:result} discusses the experimental results. Concluding remarks are presented in Section \ref{sec:conclusion}.

\section{Related Work}
\label{sec:related}

Image and video understanding has recently gained a lot of attention in deep learning research. Image classification \cite{krizhevsky2012imagenet, simonyan2014very, He2015}, object detection \cite{renNIPS15fasterrcnn,   redmon2015you}, semantic segmentation \cite{long2015fully, chen2016deeplab}, image captioning \cite{donahue2015long, xu2015show}, and localized image description \cite{densecap} tasks have witnessed tremendous progress in the last few years.

Most machine learning algorithms require inputs to be represented by fixed-length feature vectors. This is a challenging task when the inputs are variable length sentences and paragraphs. Many studies have addressed this problem. For example, \cite{kiros2015skip} presented a sentence vector representation while \cite{le2014distributed} created a paragraph vector representation. An application of such representations is shown by \cite{choi2017textually} that has used individual sentence embeddings from a paragraph to search for relevant video segments. An alternate approach uses an encoder-decoder \cite{sutskever2014sequence} framework that encodes the inputs, one at a time to the RNN-based architecture. Such an approach is shown for video captioning tasks by S2VT \cite{venugopalan2015sequence} that encodes the entire video, then decodes one word at a time. \cite{zhang2017semantic} encodes sentences with common semantic information to similar vector representations. This latent representation of sentences has been shown useful for sentence paraphrasing and document summarization.

Deep convolutional networks have shown remarkable capability to generate images. Goodfellow \textit{et al.} \cite{goodfellow2014generative} proposed an easy and effective framework of generative models based on an adversarial process. This process involves two models: the generator that captures the data distribution to generate a fake image; and the discriminator that estimates the probability of a sample being fake or real. Radford \cite{radford2015unsupervised} proposed a image generative network that was trained in an unsupervised manner and presented a set of guidelines for training the generative networks. Nguyen \textit{et al.} \cite{nguyen2016synthesizing} deals with the concept of activation maximization of a neuron and explored how each layer captures varied information using encoder, generator and visualizing networks. Other recent works expand this idea into different types of multimedia. Reed \textit{et al.}\cite{reed2016generative} explored the conversion between text to image space. Specifically, the text is encoded into a vector and consequently fed as input into the generator. Introducing an additional prior on the latent code, Plug and Play Generative Networks (PPGN) \cite{nguyen2016plug} drew a wide range of image types and a conditioner that tells the generator what to draw. Instead of conditioning on classes, the generated images were conditioned on text by attaching a recurrent, image-captioning network to the output layer of the generator, and performing similar iterative sampling.  Our work borrows concepts from such captioning and generative models to form a common vector connection space.

The notion of a space where similar points are close to each other is a key principle of metric learning. The representations obtained from this formulation need to generalize well when the test data has unseen labels. Thus models based on metric learning have been used extensively in the domain of face verification \cite{schroff2015facenet}, image retrieval \cite{gordo2016deep}, person-re-identification \cite{hermans2017defense} and zero shot learning \cite{socher2013zero}.  \cite{ngiam2011multimodal} used an auto-encoder model to learn cross modal representations and show results with audio and video datasets. Recently, \cite{wu2017starspace} leveraged this concept to associate data from different modalities. Our work can be seen as an extension of this as we extend it to visual data while linking image and caption spaces to improve image generation.

\section{Methodology}
\label{sec:method}

\begin{figure}[!ht]
  \centering
    \includegraphics[width=0.48\textwidth]{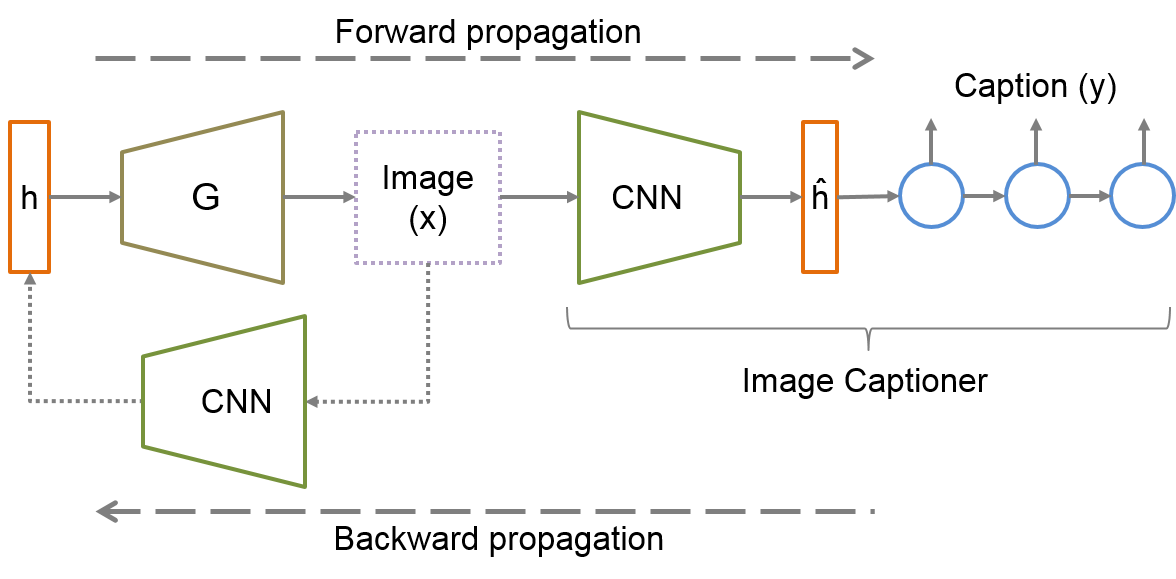}
  \caption{Caption conditioned image generation model. The image generator, G and the CNN are pretrained and are fixed. During the forward pass, we start with a random vector $h$ which generates an image (x). The image is passed through the captioner to generate a caption. The generated caption is compared with the ground truth caption (y) and the error is back propagated to update $h$ without updating the parameters in the CNN or G. After the update iterations, the image (x) is the resulting image generated using the final vector $h$.}
  \label{ppgn}
\end{figure}

Inspired by Plug and Play Generative Networks \cite{nguyen2016plug}, which reconstructs an image from a high-level feature space extracted from a pretrained encoder, we propose a similar architecture that conditions image generation on captions as shown in Figure \ref{ppgn}. This model is comprised of three pretrained modules: the generator $G$, the CNN encoder and a image captioner model. Given an image $x$, a CNN encoder is used to extract the feature vector $\hat{h}$ which is subsequently decoded by the captioner to generate a text description of the image. The reconstructed vector $h$ lies in the vector connection space, which can be used to generate another image through the pretrained generator $G$. Representation $h$ is updated based on the loss between a ground truth and generated sentence from the captioner. The loss from the captioner is back-propagated all the way back to $h$, forcing it to be representative of the semantics of the sentence. In an iterative fashion, the updated vector $h$ is passed into the generative model and the process repeats. The parameters of the encoder CNN and the generator (G) are fixed and not updated during the iterative update process. 

Equation (\ref{eqn:update}) describes the update rule for $h$ during inference. The $\gamma_1$ term represents the BLEU metric \cite{papineni2002bleu} loss associated with the generated and ground truth text. Specifically, we compute word level loss and scale it with the BLEU-1 score between the generated and ground truth text. This loss is back-propagated through the captioner and generator path to $h$. The $\gamma_2$ term is an image reconstruction loss calculated as a Euclidean distance between the pixel values of the generated and ground truth image. The $\gamma_3$ term is a vector $h$ reconstruction loss calculated as a Euclidean distance between $h$ and encoded $h$ as shown with dotted line in Figure \ref{ppgn}. The $\gamma_4$ term adds noise to the update rule. The update rule is generalizable to both \textit{image-to-vector} and \textit{sentence-to-vector} cases since the loss terms from captioner and the image reconstruction can be used in cases when ground truth text, images or both are available.

\begin{dmath}
\label{eqn:update}
h_{t} = h_{t-1} + \gamma_{1}\frac{\partial \mathcal{W}(C_{pred}, C_{gt})}{\partial h_{t-1}} + \gamma_{2}\mathcal{L}(\hat{x}, x) + \gamma_{3}\mathcal{L}(h, CNN(x)) + \mathcal{N}(0,\gamma_{4}^2)
\end{dmath}

\noindent
Where, $C_{pred}$ and $C_{gt}$ are predicted and ground truth text and $n$ is the length of the ground truth text, $\mathcal{L}$ is Euclidean distance and $\mathcal{W}$ is the word level caption loss, $\gamma_1 = BLEU(C_{pred}, C_{gt})/n$ is the scaling factor for the word loss. $\gamma_2$, $\gamma_3$ and $\gamma_4$ are hyper-parameters.

To have a better generalization for caption conditioning, we use multiple copies of the ground truth caption. This is done by using using a pretrained sentence-to-sentence model by inputting the reference caption and synthesizing paraphrase sentences. The generator is conditioned on this collection of synthesized sentences. We train a sequence-to-sequence model \cite{sutskever2014sequence} for this task as described in the next section.

\subsection{Synthesizing Artificial Sentences}

\begin{figure}[!ht]
  \centering
    \includegraphics[width=0.48\textwidth]{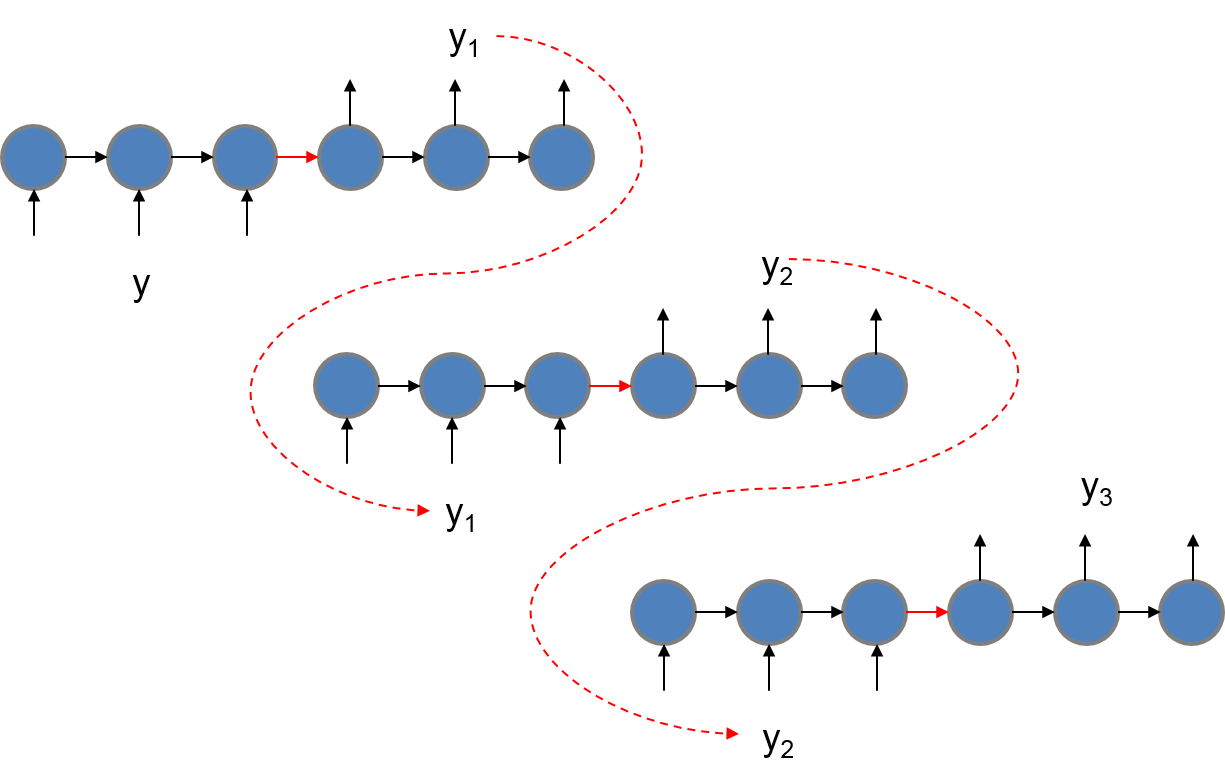}
  \caption{Synthesizing multiple sentences using a sent2sent model. Given an input sentence $y$, a pre-trained sequence to sequence model trained on caption paraphrases is used to synthesize multiple copies of $y$ with the same meaning. The input sentence $y$ generates sentence $y_1$, $y_1$ generates $y_2$, and so on.}
  \label{sent2sent}
\end{figure}

Given a reference sentence, the objective is to produce a semantically related sentence as shown in Figure \ref{sent2sent}. Recent advances at vectorizing sentences represent exact sentences faithfully \cite{kalchbrenner2014convolutional,le2014distributed,zhao2015self}, or pair a current sentence with prior and next sentence \cite{kiros2015skip}.  Just like word2vec and GloVe map words of similar meaning close to one another, we desire a method to map sentences of similar meaning close to one another. For synthesizing sentences, we consider the sentence paraphrasing framework as an encoder-decoder model.  Given a sentence, the encoder maps the sentence into a vector (sent2vec) and this vector is fed into the decoder to produce a paraphrase sentence. 

For the paraphrasing model, we represent the paraphrase sentence pairs as $(S_m, S_n)$. Let $s_m$ denote the word embedding for sentence $S_m$; and $s_n$ denote the word embedding for sentence $S_n$. $S_m$ $\in$ \{$s_1$...$s_M$\}, $S_n$ $\in$ \{$s_1$...$s_N$\} where $M$ and $N$ are the length of the paraphrase sentences. As shown in Figure \ref{sent2sent}, the input sentence $y$ generates sentence $y_1$, $y_1$ generates $y_2$, and so on.  In our model, we use an RNN encoder with LSTM cells. Specifically, the words in $s_i$ are converted into token IDs and then embedded using GloVe \cite{pennington2014glove}. To encode a sentence, the embedded words are iteratively processed by the LSTM cell \cite{sutskever2014sequence}.

The paraphrasing model was trained on visual caption datasets. There are numerous datasets with multiple captions for images or videos. For example, MSR-VTT dataset \cite{xu2016msr} is comprised of 10,000 videos with 20 sentences each describing the videos. The 20 sentences are paraphrases since all the sentences are describing the same visual input. We form pairs of these sentences to create input-target samples. Likewise, MSVD \cite{chen2011collecting}, MS-COCO \cite{Lin2014}, and Flickr-30k \cite{young2014image} are used. Table \ref{dataset} lists the statistics of datasets used.

\begin{table}[!ht]
\centering
\caption{Sentence pairs statistics in captioning datasets.}
\label{dataset}
\begin{tabular}{c|cccc}
\hline\hline
             & MSVD     & MSRVTT & MS-COCO    & Flickr  \\ \hline
\#sent       & 80K   & 200K & 123K    & 158K \\
\#sent/samp. & $\sim$42 & 20      & 5          & 5       \\
\# sent pairs& 3.2 M    & 3.8 M   & 2.4 M      & 600 K   \\
\hline \hline
\end{tabular}
\end{table}

\section{Results and Discussion}
\label{sec:result}

\begin{figure}[!ht]
  \centering
    \includegraphics[width=0.5\textwidth]{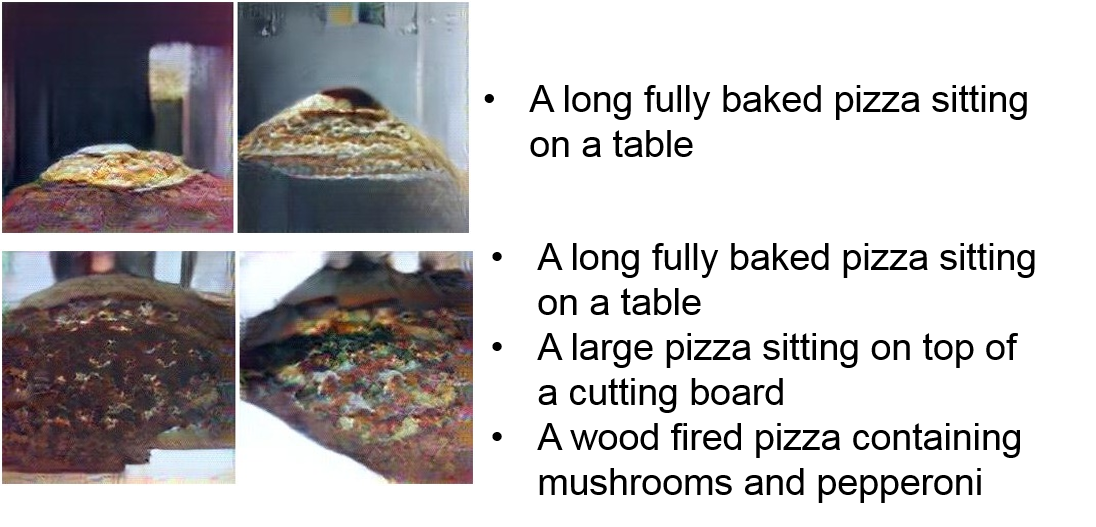}
  \caption{Image generation example conditioned on multiple captions. Top row are images generated and the corresponding input caption and bottom row are images and the corresponding three input captions.}
\end{figure}

\subsection{MS-COCO sentences}
Each image in the MS-COCO dataset has five human generated captions. We test the effect of using $>1$ captions on image generation by randomly selecting three captions among the five captions from the dataset. Punctuations are removed and the sentences are lemmatized in order to eliminate out of vocabulary words. Word gradients are averaged across all three sentences and back-propagated back to the latent vector. This combined update was observed to have stronger impact on the image quality than using a single sentence. The results are shown in Figure 3. Using multiple sentences, the generator was able to generate an image with more details as shown in the pizza image example. One explanation is that multiple captions have more objects and attributes which the generator could attend to, hence improving the resulting image quality.
Figures 4 and 5 show sample results of using three and five sentences respectively.  With respect to Figure 5, we note the generator was trained on the ImageNet dataset.  the ImageNet dataset does not have a ``person'' category. This makes it highly challenging for generating images with such categories as also noted in \cite{nguyen2016plug}. We observed that while conditioning on multiple captions, the generator was able to generate relevant images with ``person'' category. It can be seen from the example that the fine details were not clearly visible but the overall structure captures the semantic meaning of the captions.

\begin{figure}[!ht]
  \centering
    \includegraphics[width=0.5\textwidth]{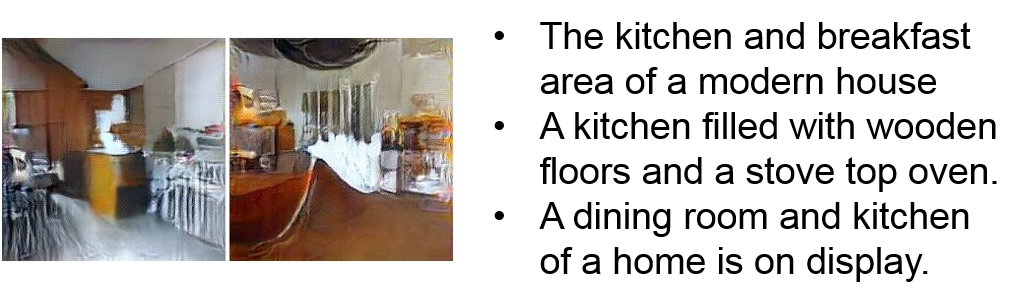}
  \caption{Image generation example conditioned on three captions from the MS-COCO image captioning dataset.}
\end{figure}

\begin{figure}[!ht]
  \centering
    \includegraphics[width=0.5\textwidth]{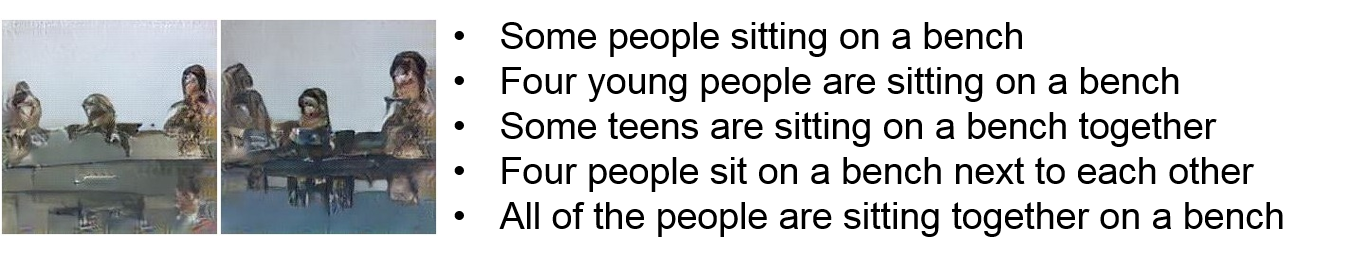}
  \caption{Image generation example conditioned on five captions from the MS-COCO image captioning dataset.}
\end{figure}

\subsection{Synthetic Sentences} 
The sentences used in the above experiment were sampled from the test set of MS-COCO and the image captioning model used in the experiment was trained on MS-COCO. In order to further validate our hypothesis, we generated synthetic sentences using a sequence-to-sequence model which produces paraphrases of an input sentence. This encoder-decoder model described in Figure 2 generates multiple paraphrase copies of an input sentence. These sentences were passed into the model for generating an image and an example is shown in Figure 6.

\subsection{Image-to-Image Analysis}
The essence of the image-to-image model can be amplified by generating a similar image given any input image. Given an input image, we pass it to an image captioning model. Multiple copies of the captions can be obtained either by using beam search in the captioner or by passing the caption through a sequence-to-sequence paraphrasing model as described earlier. Since this collection of captions capture the semantic meaning of the original image, conditioning on the captions generates images which have similar information as the original image. Different captioning models can be plugged in to compare the effect of varying generator and captioner architectures. The results are shown in Fig 6. The generator was able to attend to the white color mentioned in the captions and the overall bus structure in the original bus image.

\begin{figure}[!ht]
  \centering
    \includegraphics[width=0.5\textwidth]{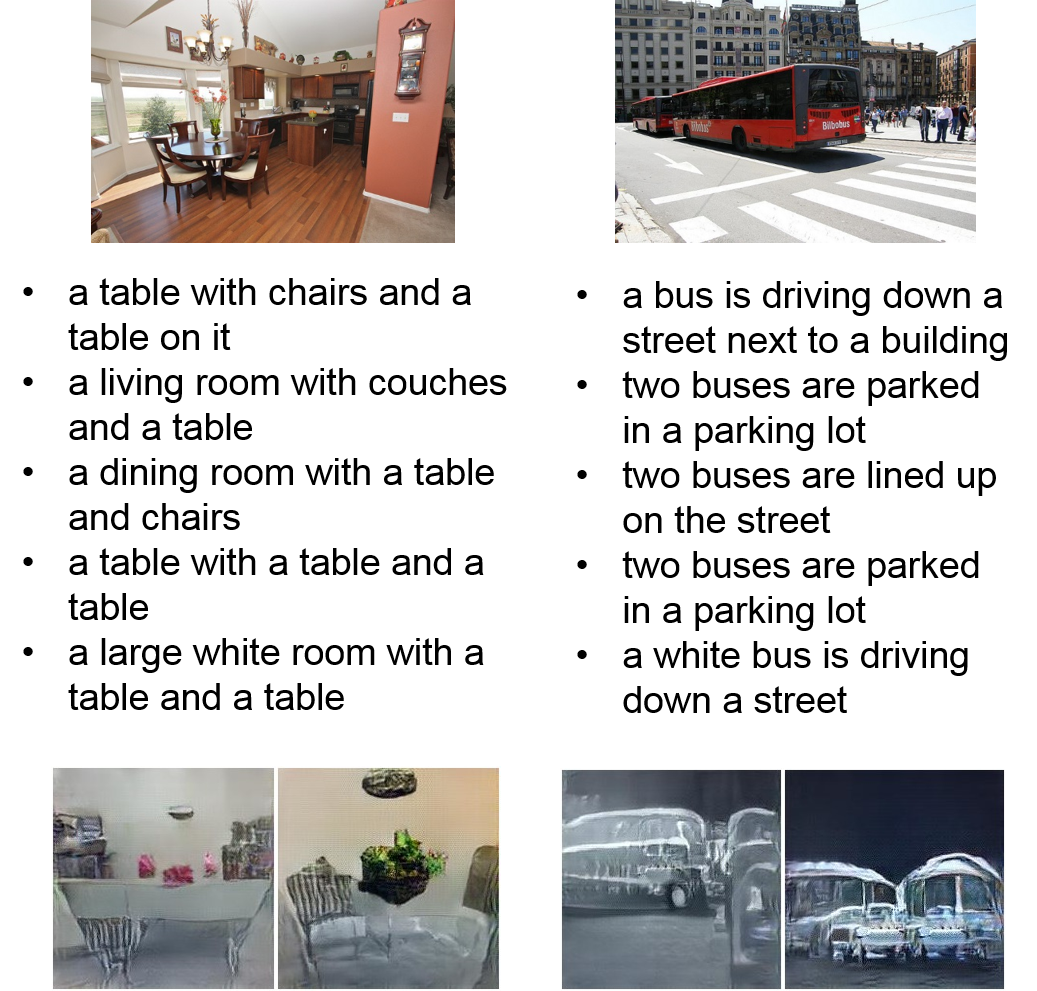}
  \caption{Image-to-image transition example. Top image is input image, center row sentences are five captions generated using the input image and bottom row images are generated images. The image generation is conditioned on multiple synthetic captions.}
\end{figure}

\subsection{Generalization}
Although back-propagating multiple word gradients did improve the search for a good quality image in the latent vector space, there were examples where we found images not representative of the captions. The effect is shown in Figure 7. Therefore, the generalization of this hypothesis is still under exploration and needs further investigation.

\begin{figure}[!ht]
  \centering
    \includegraphics[width=0.5\textwidth]{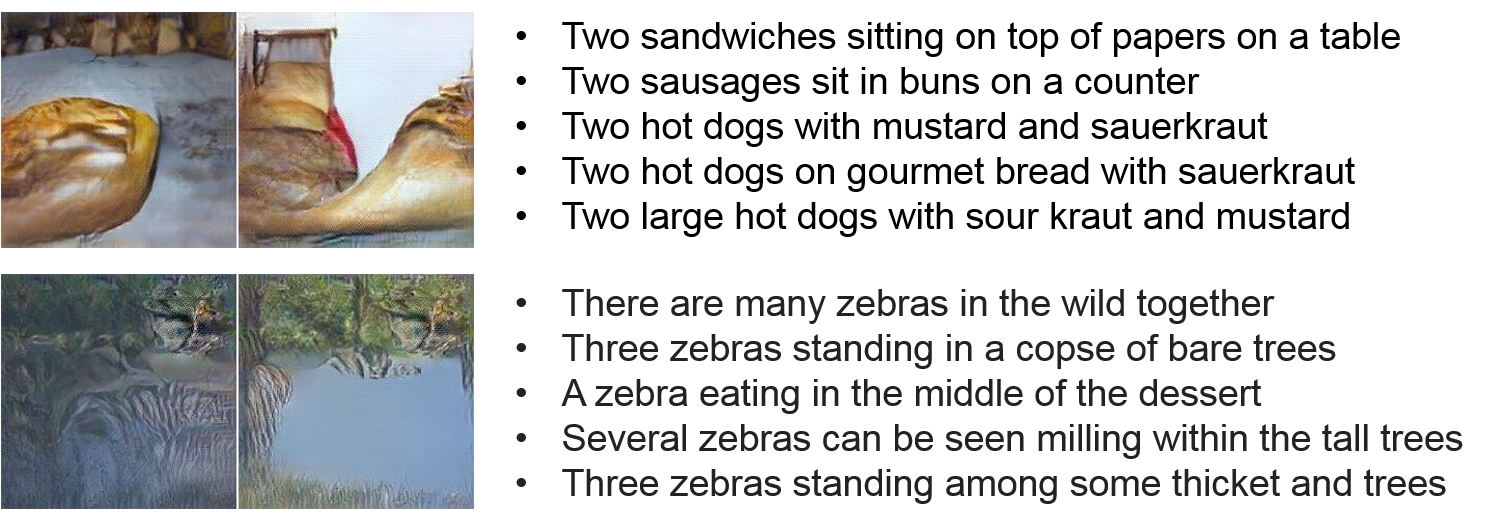}
  \caption{Some examples for which multiple captions do not help in improving the image quality.}
\end{figure}

\section{Conclusion}
\label{sec:conclusion}
This work advances the area of caption conditioned image generation by allowing image generation being conditioned on multiple sentences. The resulting vector also helps achieve a common vector space to be shared between vision and language representations. We conclude that iteratively sampling over multiple sentences indeed helps in improving the quality of the generated images. The model shows the robustness in performing cross-modal captioning. An extension of this work would be observing each layer of the generator during back-propagation. This would give more insight into the shapes and colors that are drawn by specific layers of the generator. Moreover, evaluation techniques of the generated images is still in nascent stage despite the area of image generative models having seen significant progress recently.

\section*{Acknowledgment}
The authors would like to thank NVIDIA for some of the GPUs used in this work.

\bibliographystyle{IEEEtran}
\bibliography{refs}

\end{document}